\documentclass[conference]{IEEEtran}
\IEEEoverridecommandlockouts
\usepackage{amsmath,amssymb,amsfonts}
\usepackage[ruled,longend]{algorithm2e}
\usepackage{algorithmic}
\usepackage{graphicx}
\usepackage{hyperref}
\usepackage{textcomp}
\usepackage{xcolor}
\usepackage{changepage}
\DeclareMathOperator*{\argmax}{arg\,max}

\usepackage{etoolbox}
\makeatletter
\patchcmd{\@makecaption}
  {\scshape}
  {}
  {}
  {}
\makeatother

\begin{document}

\title{Measuring Effectiveness of Video Advertisements\\
}

\author{\IEEEauthorblockN{James Hahn}
\IEEEauthorblockA{\textit{Computer Science Department} \\
\textit{University of Pittsburgh}\\
Pittsburgh, PA, USA \\
jrh160@pitt.edu}
\and
\IEEEauthorblockN{Dr. Adriana Kovashka}
\IEEEauthorblockA{\textit{Computer Science Department} \\
\textit{University of Pittsburgh}\\
Pittsburgh, PA, USA \\
kovashka@cs.pitt.edu}
}

\maketitle

\begin{abstract}
Tokyo's Ginza district recently rose to popularity due to its upscale shops and constant onslaught of advertisements to pedestrians. Advertisements arise in other mediums as well. For example, they help popular streaming services, such as Spotify, Hulu, and YouTube TV gather significant streams of revenue to reduce the cost of monthly subscriptions for consumers. Ads provide an additional source of money for companies and entire industries to allocate resources toward alternative business motives. They are attractive to companies and nearly unavoidable for consumers.

One challenge for advertisers is examining a advertisement's effectiveness or usefulness in conveying a message to their targeted demographics. A company saves time and money if they know the advertisement's message effectively transfers to its audience. This challenge proves more significant in video advertisements and is important to impacting a billion-dollar industry.

This paper explores the combination of human-annotated features and common video processing techniques to predict effectiveness ratings of advertisements collected from YouTube.  We present the first results on predicting ad effectiveness on this dataset, with binary classification accuracy of 84\%.
\end{abstract}


\section{Introduction}
In modern society, advertisements touch nearly every aspect of a person's everyday living.  Advertisements appear on television, the internet, and mobile videogames.  They are also wildly popular in videos, busy city centers, radio, billboards, and posters.  Behind every advertisement, there are content, market research, and ads research teams ensuring their company's advertisements achieve the highest click rate possible, converting the audience into active consumers.  In general, when creating advertisements, the content and evoked emotion are heavily taken into consideration.  For example, Ford wants to develop commercials with cars.  Ford commercials generally want to evoke emotions to ensure their consumers feel `safe' or `positive'.  In contrast, a company developing home security systems wants to evoke `fear' into the audience in the sense that consumers are fearful of burglars, so they must act fast and buy a security system.  Clearly, advertisements are multifaceted creations with the ability to convey symbolism, propaganda, emotions, and careful thinking.

In particular, when designing advertisements, the main goal is ensuring effectiveness.  Generally, an effective ad implies a higher conversion rate for consumers buying a product.  Effectiveness can be formulated by delivering a product and message in a unique, straightforward manner that stands out to consumers.  The advertisements industry's market cap was over \$200 billion in 2018 \cite{c27}.  In fact, in 2017, Fox charged \$5 million for a 30 second ad during Super Bowl LII, reaching more than 100 million viewers \cite{c26}.  Thus, even a trivial boost in an ad's effectiveness can lead to a millions of dollars of added revenue.  One rising problem with current advertisements is their failure to draw in millennials \cite{c11}.  Therefore, the investigation of advertisement effectiveness is not only important to boost revenue and prestige of a company's products, but to also engage the next generation of consumers.  In fact, Hulu is implementing ad selectors \cite{c28}.  These ad selectors allow viewers to select one of many displayed ads, therefore handing some of the power to the consumer, rather than force feeding generalized content to large audiences.  One research paper from 2012 investigated a similar idea to ad selectors, the video ad 'skip' feature on YouTube videos \cite{c22}.

\begin{figure}[t]
  \centering
  {\label{ref_label1}\includegraphics[width=250px,keepaspectratio]{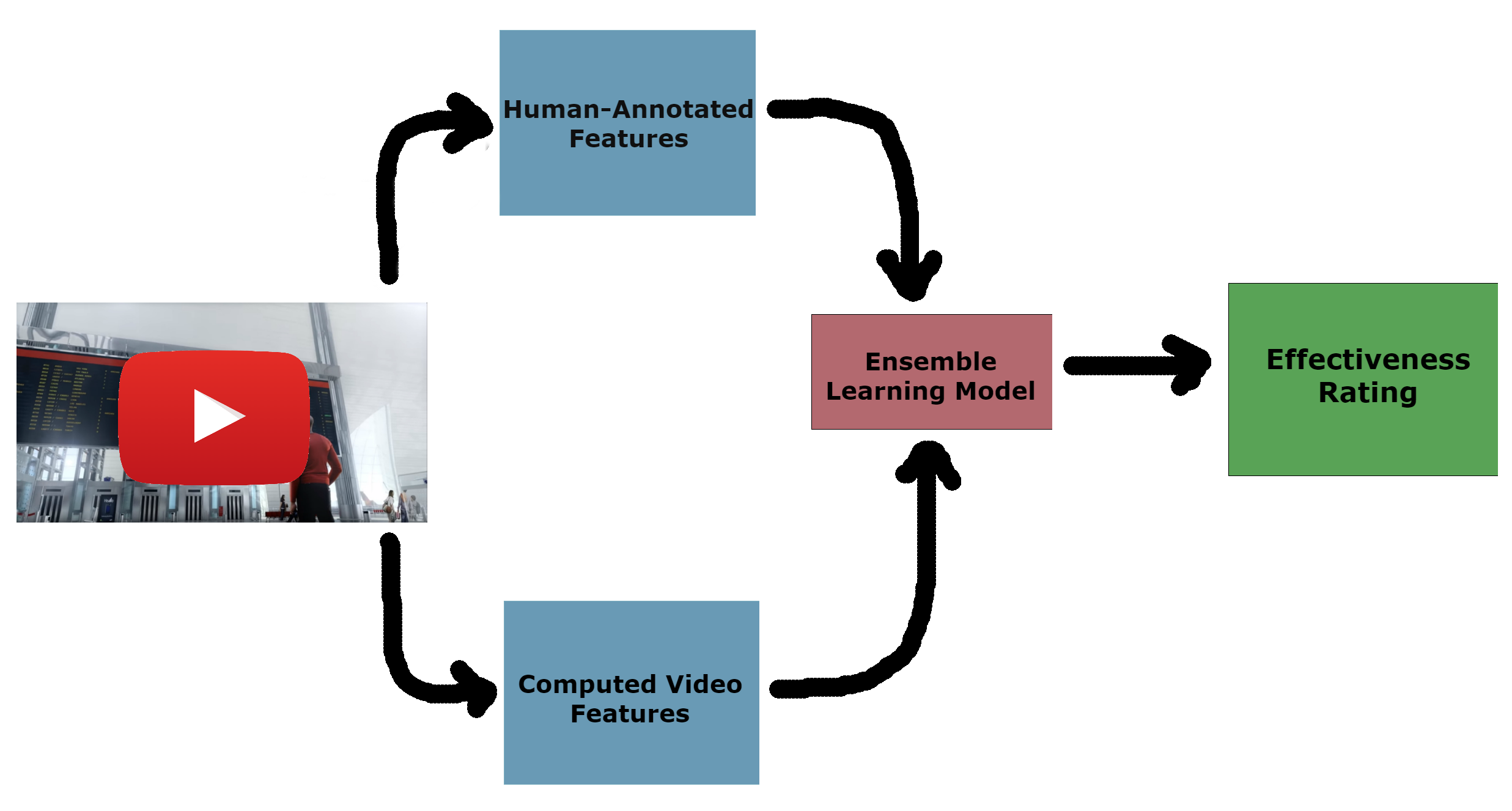}}
  \caption{Proposed model for learning video advertisement effectiveness}
  \label{learning-model}
\end{figure}

In this paper, we investigate a combination of features gathered from human annotators, common video processing techniques, and features extracted by Ye et al.  Some of these features are low-level computer vision, such as shot boundary, average hue, and optical flow, others are higher level, such as object detection, facial expression detection, and text, while a few encapsulate higher level information about the advertisement, such as memorability, climaxes, and duration.  A thorough investigation is performed to gather inference on the distribution and approaches to investigating this challenging task, while preprocessing any features and the dataset beforehand.  Next, support vector machines, decision trees, and a logistic regression classifier are trained on the features, where one classifier is trained on one feature at a time.  Individual classifiers performed near the baseline on the binary, four-way, and five-way classification tasks.  However, a unique, hybrid ensemble learning algorithm is introduced to model the underlying distribution of the training dataset, which is used to predict effectiveness ratings of the testing dataset, resulting in significantly higher accuracies on all three tasks, ranging from a boost of 30-35\%.  As such, this paper utilizes a mix of human-annotated labels and computational video processing techniques to gather insights into effectiveness for advertisements, unseen in this field since most studies involve human surveys or a computational approach, but from a human observation standpoint.  This process requires a trivial amount of input to predict any video's effectiveness on any media platform, any resolution, any type of advertisement, and in a non-specific culture.  Thus, this paper presents a first contribution to automated, human-less video effectiveness contribution and inference, as shown in Figure \ref{learning-model}.

\section{Related Work}
The results presented in this paper are the first of their kind in terms of measuring advertisement effectiveness on this dataset.  \cite{c40} examines video effectiveness using gaze detection, a computer vision approach, but this data is not always readily available or easily extracted from a video without further user studies and exposure to the video.  \cite{c41}, from CVPR 2018, is the closest paper to our research, both examining video effectiveness and performing analysis using computer vision techniques.  Despite this, the experiments differed significantly, as Okada et al. measured gaze detection, facial expressions, heart rate, and ``physiological responses'' by processing videos of Japanese people's reactions while watching the advertisements.  As such, they limit the human population to a demographic of Japanese civilians and utilize computational procedures to extract physiological reactions from patrons watching videos, rather than video extracted features and human-annotated labels.  Additionally, \cite{c42} predicts click-through rate as a measure of effectiveness for advertisements.  While applicable on many devices, this metric does not apply to television commercials and public displays.  Also, click-through rates may not always be the best determinant of effectiveness.  For example, public service announcements, network content ads, and pop-up video ads on websites generally do not contain or hyperlink or metric to monitor clicks and click-through rates are not indicitive of the time spent on the website and how many pages the user visits, limiting their paper's approach.  Therefore, our dataset and approach is more robust and covers a wider variety of advertisements.  Finally, this problem is two-fold: it contains the technical challenges in computer vision and the arts/media challenges in advertisements.

To our knowledge, much of the advertisements research is contained specifically in advertising journals, such as the Journal of Advertising Research from the Advertising Research Foundation \cite{c29} and the Journal of Advertising from the American Academy of Advertising \cite{c30}.  As such, previous research is mostly contained in media studies, rather than the technological industry.

In terms of work related to advertisement effectiveness, previous media arts papers have focused on emotion-induced engagement [25], reduced effectiveness of sexually suggestive advertisements \cite{c23}, understanding context of an ad to boost effectiveness \cite{c21}, and analysis of ad content and repitition to predict effectiveness for TV commercials \cite{c16}.  Other papers discussed brand recognition \cite{c20}, intellectual arousal and effectiveness \cite{c19}, and analysis of internet banner ad click-through rate to measure effectiveness \cite{c18}.  William investigated measuring ad effectiveness, exploring the concept before much of the digital age \cite{c17}.  Recently, Forbes completed a survey asking readers to identify key concepts in an effective or memorable ad \cite{c15}.  Responses included humor, a tagline, jingle, iconic characters, thought-provoking content, and rapport of trust and loyalty.  One response surprisingly claimed hurting people creates a memorable ad.  Determining features of effective videos is clearly a difficult challenge, as ``thought-provoking content'' is an abstract content and is hard to convert into a computable form.  \cite{c14} analyzed video advertisement effectiveness on a dataset from Akamai, but included features such as the content provider (i.e. sports, movies, news, etc.), and the position of the ad within the content (i.e. pre-roll, mid-roll, and post-roll).  Additionally, they considered specific demographics of individual viewers, as well as the final result of a video's views and impressions to determine effectiveness.  \cite{c44} contrasts a person's personality and ad effectiveness, which is not the aim of this paper.  \cite{c45} surveys cross-media ad effectiveness by analyzing mobile and desktop ads for brand impact, focusing moreso on platforms for effectiveness.  Finally, Readex Research performs ad effectiveness studies, but they send out surveys to human audiences to collect opinion for each advertisement, which can be a long process, and they do not involve the computational aspect that this paper presents \cite{c46}.  While these papers revolved around advertisements, media studies, and ad effectiveness, which are closely strung to this paper, they have three significant differences.  First, most of these papers require human input from surveys, which requires money and time to find participants and aggregate results.  Second, most of the studies do not perform computational analysis or prediction of effectiveness, so the process is not automated and analysts must investigate each feature individually to perform predictions.  Finally, even if the paper involved computation and learning algorithms, they were niche in terms of brands and products, cultures of participants, or features, limiting the scope of the paper, where an increase in one feature might weaken the impact of another feature, which is a serious concern since advertisements are an agglomeration of key features.

In this paper, effectiveness is measured by human-annotated labels, so results directly relate to whether humans believe a specific ad is effective.  A given video by itself may have million of views, but it can be polarizing for its audience.  For example, Justin Bieber's ``Baby'' music video on YouTube \cite{c32} has two billion views, but was the most disliked video on the website for years with ten million dislikes to its ten million likes.  Oddly enough, YouTube's very own ``YouTube Rewind 2018: Everyone Controls Rewind'' video has 160m views, 2.4m likes, and 15m dislikes \cite{c33}.  Clearly, views and brand are not always an accurate measure of advertisement effectiveness.

Finally, in 2006, researchers investigated the use of neural networks to predict television ad effectiveness \cite{c24}.  They achieved an accuracy of 99\% on a 10-scale effectiveness utilizing 50 features and a dataset of 837 respondents, but the study possesses numerous tight constraints that allow for lower variability of responses, contributing to such a high accuracy.  Therefore, \cite{c24} solved a problem similar to the one in this paper.  However, there are four key differences.  First, and perhaps most significant, all 837 participants in the survey analyzed one of three ads, all of which were marketing toothpaste.  Our dataset consists of several magnitudes more topics and product types, thus making the task more challenging.  Second, their dataset only considered television ads, while Hussain et al. considers internet, television, and general media ads \cite{c1}.  Third, this dataset was collected in 2006, before the rise of mass digital adoption.  Mobile phones were barely popular around the time, resulting in less modernity of their dataset.  Fourth, all participants were from India in a university setting.  Hussain et al.'s dataset consists of random sampling of citizens through the use of Amazon Mechanical Turk, thus giving a well-rounded background of demographics.  Therefore, this paper's research is considerably different from any prior research due to the dataset novelty, technological application, and success despite high variety of data.  It provides the first mix of media studies with state-of-the-art computer vision technological approaches to predict internet and television advertisement effectiveness with a diverse audience and video dataset.

\section{Methods}

First, we will discuss the dataset, along with its features, data collection procedure, and brief overview into the values they can hold.  Next, preprocessing of the data will be discussed in terms of removing class imbalance.  Then, the discussion transitions to additional features extracted from the advertisements and features collected from Ye et al.'s dataset, as well as their importance.  Fourth, this section contains exploratory analysis on the advertisements dataset, showcasing topics and sentiments distributions, effectiveness distributions for topics and sentiments, correlations between features and effectiveness, and reliability of the human annotators' ratings across the dataset.  Also, we will discuss the learning methods through the use of support vector machines, decision trees, and logistic regression classifiers, as well as an ensemble of classifiers.  Finally, the experiments section will showcase the key results of the learning process and wrap up any notable conclusions about this dataset.

\subsection{Dataset}
Hussain et. al released a dataset to CVPR 2017 collected from Amazon Mechanical Turk human annotators.  The dataset consists of two parts: 64,832 static image advertisements and 3,477 video advertisements.  Both datasets are similar with few differences in features.  The static image dataset possesses labels for the topic, sentiment, question/answer statement, symbolism, strategy, and slogan.  In comparison, the video dataset has labels for topic, sentiment, action/reason statement, funniness, degree of excitement, language, and effectiveness.  

Topics are in a range of thirty-eight possibilities describing the overall theme of the advertisement, such as `cars and automobiles', `safety', `shopping', or `domestic violence'.  Sentiments describe emotion evoked in the user, such as `cheerful', `jealous', `disturbed', `sad', and more, with thirty possibilities.  Funniness and excitement are binary variables with value 0 indicating unfunny/unexciting and 1 implying funny/exciting.  The ternary language feature takes the value 0 if the advertisement is non-English, 1 if it English, and -1 if language is unimportant to the video, such as a voiceless ad.  Action/reason statements consist of a simple call to action and motivation statement combined into one.  For example, one automobile commercial uses the action/reason statement, ``I should buy this car \textit{because} it is pet-friendly.''  Every statement's action and reason are broken up with 'because'.  As such, the action asked of the consumer it to buy the car and the reason is its pet-friendly characteristic.  Action/reason statements vary in complexity throughout the dataset.  Finally, the goal of this research is to predict the output label, which is the 'effective' feature.  Effectiveness is also a discrete value ranging from one to five.

All videos were gathered from YouTube and verified as an advertisement rather than an unrelated video.  Then, human annotators on Amazon Mechanical Turk labelled all seven features for each video.  Five annotators were assigned to each video to control for possible high variance in labels but were kept anonymous.  Therefore, controlling for bias in work identity is unfeasible in these experiments.  Additionally, the raw version of the video dataset contains all ratings for each feature for each video, while an alternative, clean version utilizes mean or mode across all five labels of each feature to compute a simplified, or 'clean', representation of that video's ratings.

\subsection{Data Preprocessing}
The most immediate issue in terms of preprocessing is ensuring class balance.  After investigation, we discovered the dataset consists of 193 samples of effectiveness 1, 261 samples of effectiveness 2, 1319 samples of effectiveness 3, 426 samples of effectiveness 4, and 1278 samples of effectiveness 5.  The overall effectiveness for a given video is computed as the mode of the five ratings for the video.  In general, any numeric, discrete feature with several human-annotated labels was aggregated by use of mode to better represent the video's underlying ground truth value.  This specifically refers to the provided clean dataset \cite{c1}, which are the effectiveness, exciting, funny, language, sentiments, and topics features.  In case of ties, the lowest value was chosen.  To ensure class balance, the class with lowest count determined the number of randomly sampled videos from each class. Therefore, 193 samples were used for each class, reducing the dataset's size to 965 and increasing difficulty of the task significantly.

\begin{figure}[t]
  \centering
  {\label{ref_label1}\includegraphics[width=231px,keepaspectratio]{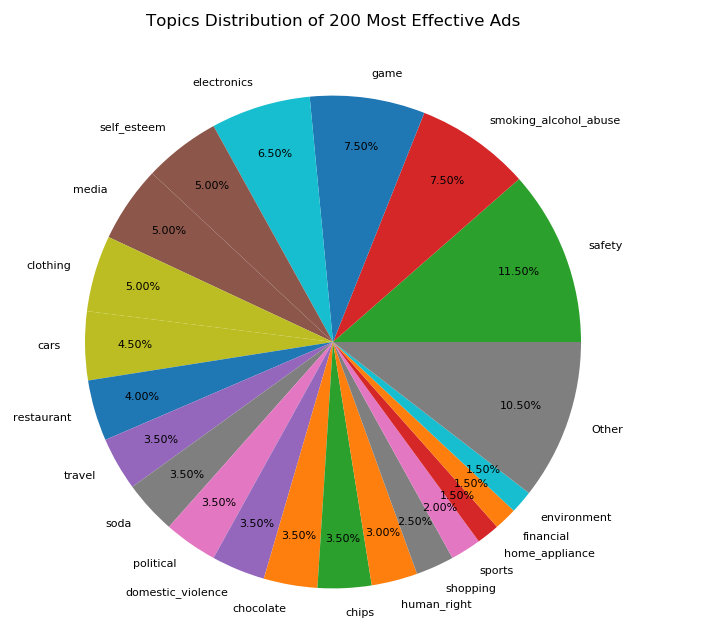}}
  {\label{ref_label2}\includegraphics[width=231px,keepaspectratio]{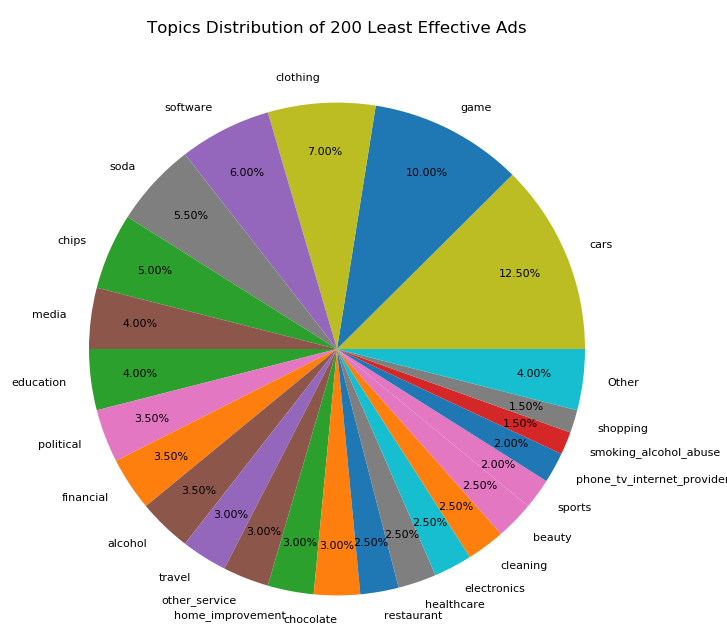}}
  \caption{Normalized distribution of topics across 200 most effective and least effective advertisements}
  \label{topics-distributions-pie}
\end{figure}

\begin{figure}[t]
  \centering
  {\label{ref_label1}\includegraphics[width=200px,keepaspectratio]{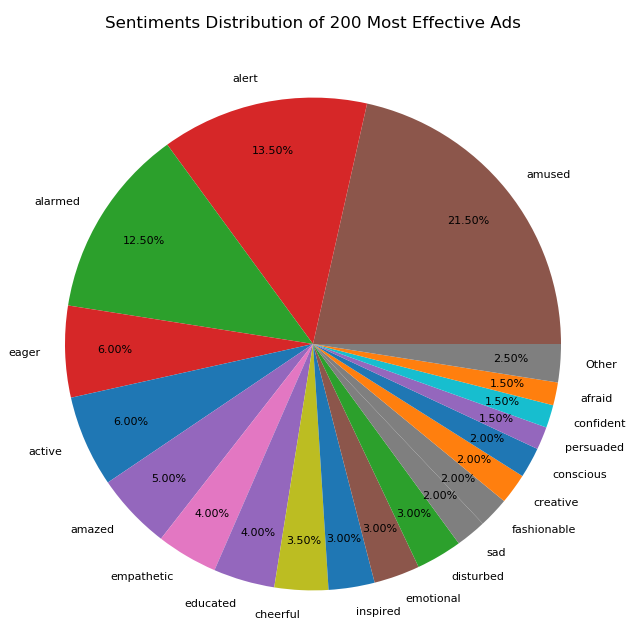}}
  {\label{ref_label2}\includegraphics[width=200px,keepaspectratio]{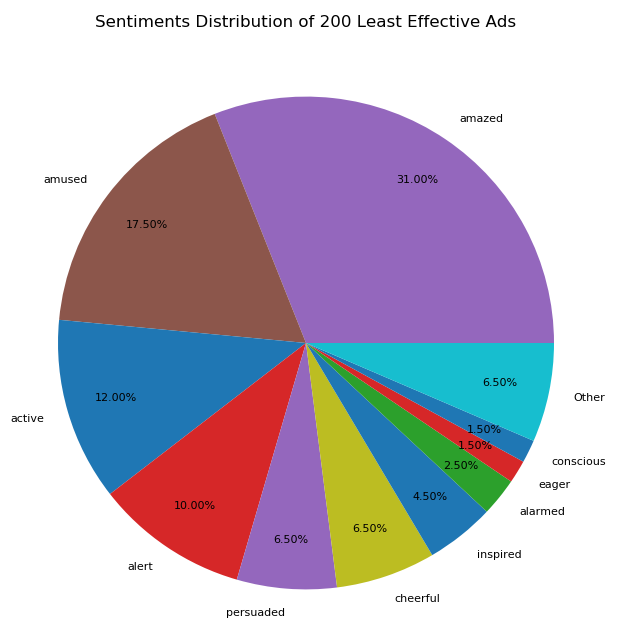}}
  \caption{Distribution of sentiments across 200 most effective and least effective advertisements}
  \label{sents-distributions-pie}
\end{figure}

\subsection{Features}

The features gathered for analysis and future learning are a combination of human-annotated labels, output from popular video processing techniques, and Ye et al.  Twenty-one features were used for the machine learning approach, while correlations were calculated for an additional six features.  As such, this subsection is broken down into two further sections exploring features from video processing techniques and data collected from \cite{c3}.  Features collected from humans are not discussed as they have been previously mentioned are straightforward in their collection; that data was already cleaned and pre-packaged with the dataset.\\

\subsubsection{Low-level and Computed Features}
\begin{itemize}
    \item Color
    \begin{itemize}
        \item Average Hue
        \item Median Hue
        \item Average Intensity
        \item Average Intensity over middle 30\%
        \item Average Intensity over middle 60\%
    \end{itemize}
    \item Average Memorability
    \item Video Duration
    \item Text
    \begin{itemize}
        \item Average Word Length
        \item Meaningful Words
        \item Average Sentence Length
        \item Most Common Word
    \end{itemize}
    \item Shot Boundaries
    \item Optical Flow
\end{itemize}
Next, to gain further insights into the factors that contribute to an ad video's effectiveness, fourteen new features were developed and computed from each video.  The first five deal with the colors and visuals: average hue, median hue, average intensity, average intensity over middle 30\% of video, and average intensity over middle 60\% of video.  Hue is classified as a 3-dimensional vector of a pixel's red, blue, and green color value.  Intensity is calculated as the greyscale value of a pixel.  The latter two features attempt to gauge the most captivating portions of the video.  The middle 30\% of a video is a window covering 30\% the size of the height and width located in the center of the image; the middle 60\% is computed in similar fashion.  Next, average memorability across frames is computed utilizing \cite{c2}.  Duration of the video is gathered from calls to the YouTube API.

Text content is gathered from Google Cloud Vision's optical character recognition (OCR) API.  The average duration of a video is 15 seconds at 24 frames per second, thus consisting of 360 frames.  Every $60^{th}$ frame is extracted from the video, totalling 6 frames of text information per video on average.  All 6 frames were passed into the API and text was extracted, resulting in four new features: average word length, meaningful words contained, average sentence length, and most common word.  A meaningful word is defined as any non-trivial word from text, such as proper nouns, locations, objects, and adjectives.  It is important to note popular text preprocessing was performed to ensure useful results, especially when finding meaningful words.  For example, each word was stemmed (i.e. 'confused'/'confusing' become 'confuse') and stop words (i.e. 'a'/'the'/'by'/etc.) were omitted to ensure meaningfulness of features.  These preprocessing techniques were handled out with the popular Python library, NLTK \cite{c13}.

Furthermore, shot boundaries were computed in addition to average optical flow of videos.  Shot boundaries were measured by counting the number of scene changes throughout the video.  This measures the video's quickness; higher scene changes equates to a faster video.  The reason for using this metric is analyzing whether fast or slower videos translate to higher or lower effectiveness due to speed of delivery of the author's message.  Additionally, optical flow is computed as the sum of the average optical flow change from frame to frame.  Therefore, it is seen as a summation of vector magnitudes, representing the change in the video's content; higher optical flow is equivalent to intense content shifts.  Finally, optical flow was converted into a 30-bin across the entire video.  Every bin consists of the sum of vector magnitudes for that portion of frames in the video.  Then, the bin was normalized via L1 norm such that all bin values sum to one.\\

\begin{table*}[ht]
    \begin{center}
    \begin{adjustwidth}{0.7in}{}
    \begin{tabular}{|l|l|l|l|l|l|}
    \hline
    \textbf{Classifier} & \textbf{Feature} & \textbf{Binary Accuracy} & \textbf{Four-way Accuracy} & \textbf{Five-way Accuracy} \\ \hline
    SVM & Topics & 0.5627 & 0.3178 & 0.2094 \\ \hline
    SVM & Sentiments & 0.5386 & 0.2940 & 0.2435 \\ \hline
    SVM & Memorability & 0.5037 & 0.2499 & 0.2072 \\ \hline
    SVM & Optical Flow & 0.5060 & 0.2508 & 0.2006 \\ \hline
    SVM & Cropped 30\% & 0.4977 & 0.2557 & 0.1977 \\ \hline
    SVM & Cropped 60\% & 0.5022 & 0.2525 & 0.2142 \\ \hline
    SVM & Average Hue & 0.5151 & 0.2966 & 0.2300 \\ \hline
    SVM & Median Hue & 0.5102 & 0.2823 & 0.2142 \\ \hline
    SVM & Duration & 0.5562 & 0.2586 & 0.2281 \\ \hline
    SVM & Text Length & 0.5027 & 0.2411 & 0.2141 \\ \hline
    SVM & Meaningful Words & 0.4909 & 0.2617 & 0.2158 \\ \hline
    SVM & Average Word Length & 0.4966 & 0.2463 & 0.1905 \\ \hline
    SVM & Word Count & 0.5064 & 0.2538 & 0.2011 \\ \hline
    SVM & Sentiment Analysis & 0.4726 & 0.2480 & 0.2187 \\ \hline
    SVM & Audio & 0.5246 & 0.2625 & 0.2195 \\ \hline
    SVM & Objects & 0.4785 & 0.2494 & 0.2033 \\ \hline
    SVM & Places & 0.5224 & 0.2676 & 0.2153 \\ \hline
    SVM & Expressions & 0.4950 & 0.2366 & 0.2107 \\ \hline
    SVM & Emotions & 0.4952 & 0.2574 & 0.2266 \\ \hline
    SVM & Climax & 0.3589 & 0.2533 & 0.2175 \\ \hline
    SVM & All Features Aggregated & 0.5669 & 0.3420 & 0.2398 \\ \hline
    SVM & All Text Features Aggregated & 0.5103 & 0.2822 & 0.2269 \\ \hline
    Decision Tree & Topics & 0.5720 & 0.3288 & 0.2284 \\ \hline
    Decision Tree & Sentiments & 0.5464 & 0.3083 & 0.2413 \\ \hline
    Logistic Regression & Exciting & 0.5595 & 0.3036 & 0.2423 \\ \hline
    \multicolumn{2}{|c|}{\textbf{Combined Ensemble}} & 0.8485 & 0.6501 & 0.5505 \\ \hline
    \multicolumn{2}{|c|}{\textbf{Baseline}} & 0.5000 & 0.2500 & 0.2000 \\ \hline
    \end{tabular}
    \end{adjustwidth}
    \end{center}
    \caption{Accuracies of individual classifiers, and the combine ensemble displayed in Algorithm \ref{ensemble-algo}, averaged over five simulations for binary, four-way, and five-way classification tasks}
    \label{accuracies}
\end{table*}

\subsubsection{Ye et al. features}
\begin{itemize}
    \item Audio
    \item Objects
    \item Places/Scenes
    \item Facial Expressions
    \item Emotions
\end{itemize}
\cite{c3} investigated climax in video advertisements.  They gathered additional features on the video dataset including facial expressions, emotions, audio, common objects, detected climaxes, and scenes.  All of these features were utilized to gain further insight into advertisement effectiveness.  

The audio signal per frame was averaged to represent a video's 'loudness'.  Furthermore, all objects contained in the videos was gathered.  Then, the probability of each object's occurrence was calculated (i.e. 1000 objects were detected, and 800 were 'person', so 'person' now has value 0.8) on the entire dataset to provide prior probabilities.  Finally, each video's individual object probability distribution was calculated and each object's probability was divided by the respective prior to calculate the final distribution.  For example, if the probability of a person in a specific video is 0.6, but the aforementioned prior is 0.8, then the final value in the feature vector is 0.6 / 0.8 = 0.75.  This indicates a person appears 75\% of the time relative to the average.  In total, 786,602 objects were discovered across 80 unique classes, representing a diverse array of objects.  There were 872,870 detected places/scenes, 365 of which are unique, 34,572 detected facial expressions, 8 of which were unique, and 3,659,717 detected emotions, 26 of which were unique.  These three aggregated features were preprocessed in similar fashion to the objects feature (i.e. a prior was calculated to determine the probability across the entire dataset and each video's value is a ratio between its probability distribution over the prior).  

The audio signal (one-dimensional), object probability distribution (80-dimensional), as well as places, expressions, and emotions distributions, were used as separate feature vectors for future training.  The number of climaxes per video was summed up to represent an additional feature to represent the amount of highlights in a video.

These features are valuable since they allow more analysis, correlations, and construction of learning models, which in turn provides more insight into the dataset.  Also, each by itself is important in analyzing most of the content placed throughout the video, such as the objects and facial expressions, which were not gathered in the previous dataset, while climax data provides a high level perspective of how action-packed the advertisement is, which can lead to the user being more engaged.  Without these features, the overall dataset is left without important content in the video, not gathered previously, that helps provide insights into the content.

\subsection{Data Analysis}
Simple data analysis was performed to view general trends of the dataset.  Since effectiveness is the output label, correlations between each feature and the output label were computed with results demonstrated in Table II.  Correlations were performed on the entire dataset.  All correlations were weak or non-existent.  The features representing number of shot boundaries and number of unique annotated sentiments performed the worst.  As such, fitting a linear regression line between any of the features and effectiveness will provide poor results and accuracy.  Despite this, the duration, exciting, and audio features showcase the strongest positive or negative correlations, so they were later selected as features to include in our ensemble.

Ensuring consistency across the human annotators and their effective ratings is key to prevent outliers.  Unfortunately, kappa statistics are not available for use on this dataset since anotators are anonymous and it is not possible to match one annotator's ratings to specific videos.  However, the coefficient of variation, represented as $c_v = \frac{\sigma}{\mu} \times 100\%$, measures volatility of a distribution, which in this case will be the ratings for a given video, as a quality assurance check \cite{c39}.  There were 3114 videos (89.56\%) with $c_v \leq 0.5$.  Additionally, there were 2932 (84.33\%) videos with $c_v \leq 0.4$ and 2244 (64.54\%) videos with $c_v \leq 0.3$.  This indicates a majority of videos were under 30\% variability in terms of ratings and about 20\% of videos had somewhere between 30\% and 40\% variability.  Clearly, the ratings of videos were relatively consistent and reliable.

\begin{table}[h]
    \begin{center}
    \begin{tabular}{|l|l|}
    \hline
    \textbf{Feature X}                & \textbf{Correlation(X, effective)} \\ \hline
    Duration                          & 0.207                     \\ \hline
    Exciting                          & 0.181                     \\ \hline
    Language                          & 0.146                     \\ \hline
    Funny                             & 0.101                     \\ \hline
    \# of Detected Climaxes           & 0.034                     \\ \hline
    \# of Unique Annotated Sentiments & 0.028                    \\ \hline
    \# of Shot Boundary Changes       & 0.026                     \\ \hline
    Entropy of Optical Flow Bins      & -0.011                    \\ \hline
    Avg. Length of Action Response    & -0.056                    \\ \hline
    Audio Signal                      & -0.071                    \\ \hline
    Avg. Length of Reason Response    & -0.117                    \\ \hline
    \end{tabular}\newline\newline
    \caption{Correlations between useful features and effectiveness \cite{c4}}
    \end{center}
    \label{correlations}
\end{table}

\begin{figure}[t]
  \centering
  {\label{ref_label1}\includegraphics[width=200px,keepaspectratio]{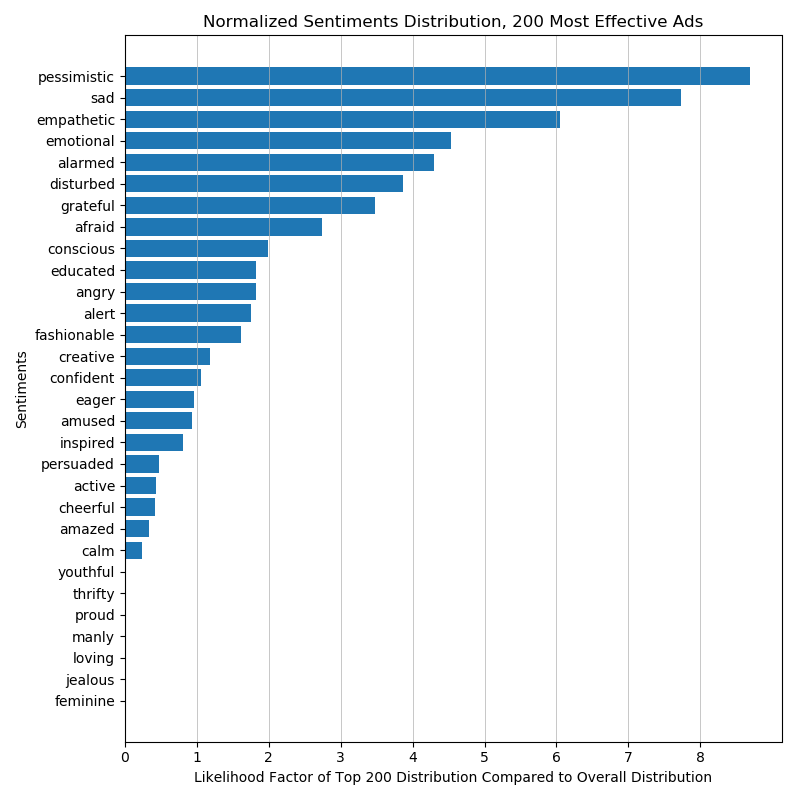}}
  {\label{ref_label2}\includegraphics[width=200px,keepaspectratio]{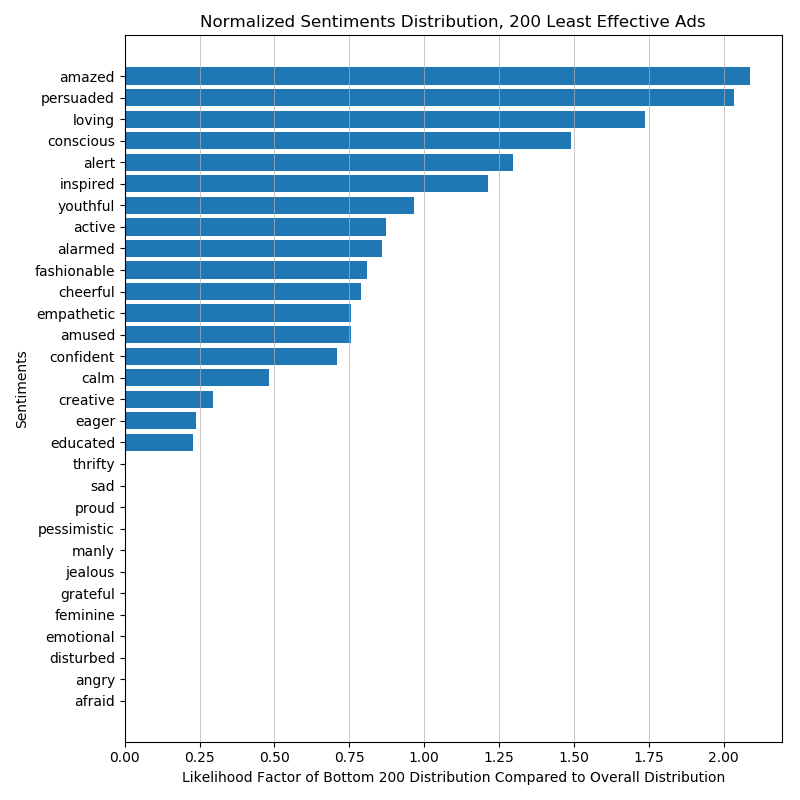}}
  \caption{Distribution of sentiments across 200 most effective and least effective advertisements}
  \label{sents-distribution-bar}
\end{figure}

\begin{figure}[t]
  \centering
  {\label{ref_label1}\includegraphics[width=200px]{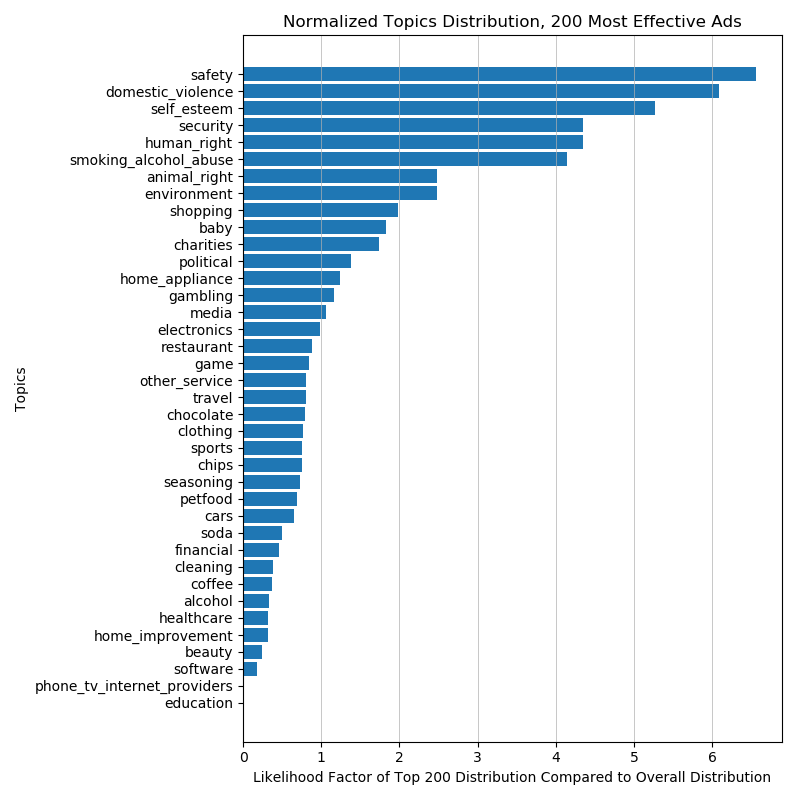}}
  {\label{ref_label2}\includegraphics[width=200px,keepaspectratio]{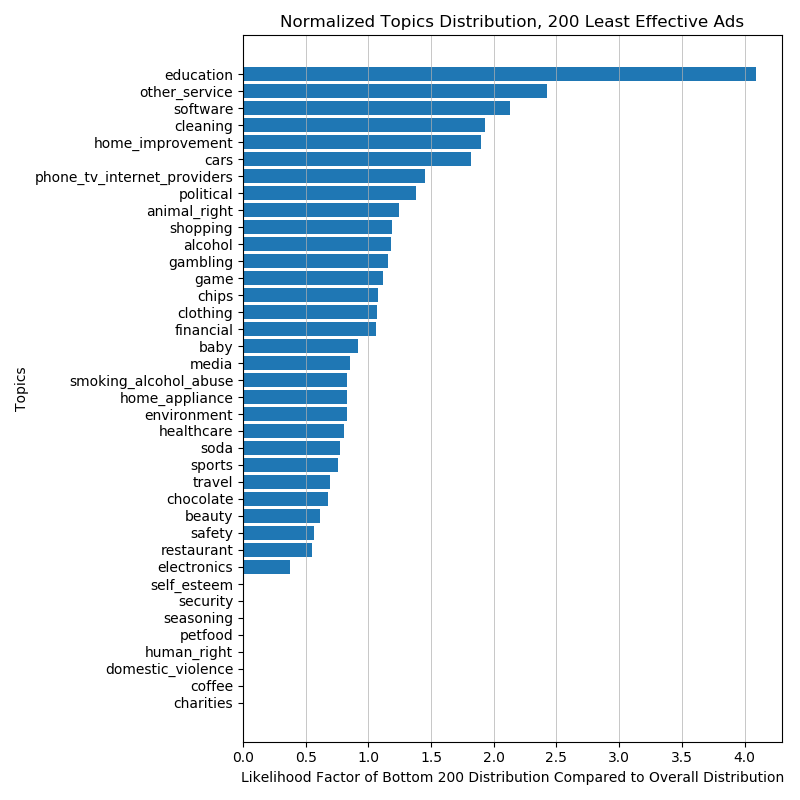}}
  \caption{Distribution of topics across 200 most effective and least effective advertisements}
  \label{topics-distribution-bar}
\end{figure}

\begin{figure}[h]
  \centering
  {\label{ref_label1}\includegraphics[width=200px,keepaspectratio]{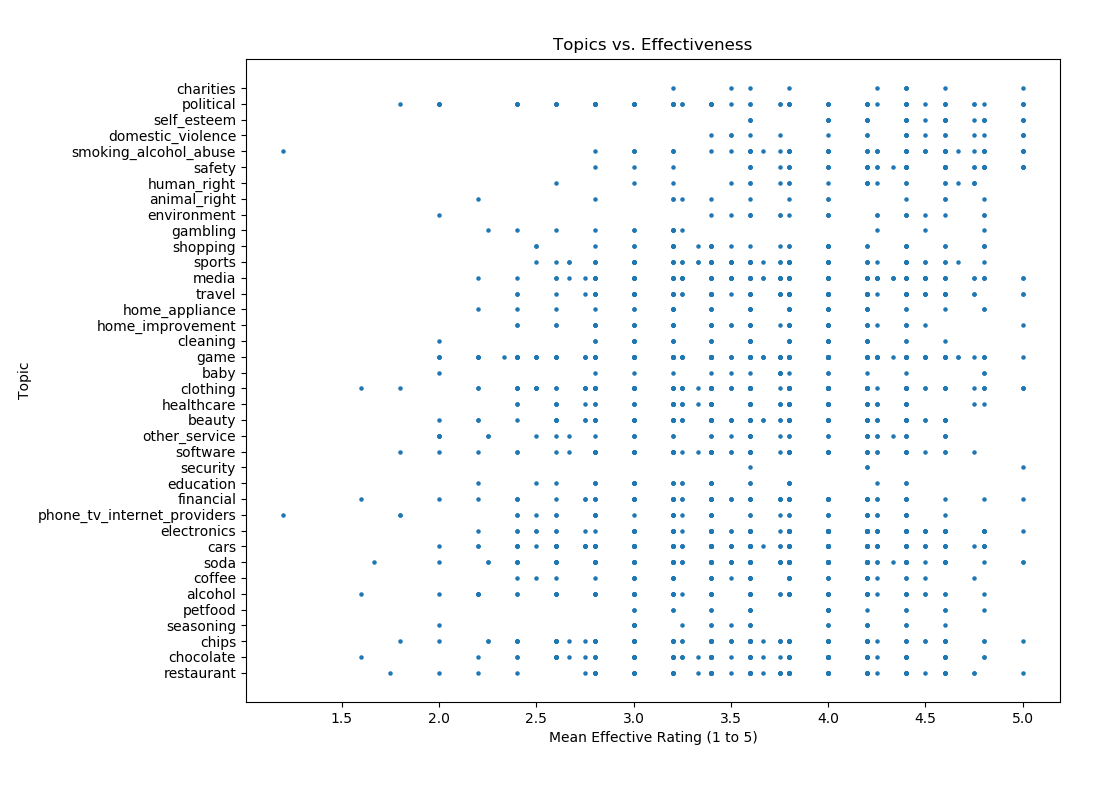}}
  {\label{ref_label2}\includegraphics[width=200px,keepaspectratio]{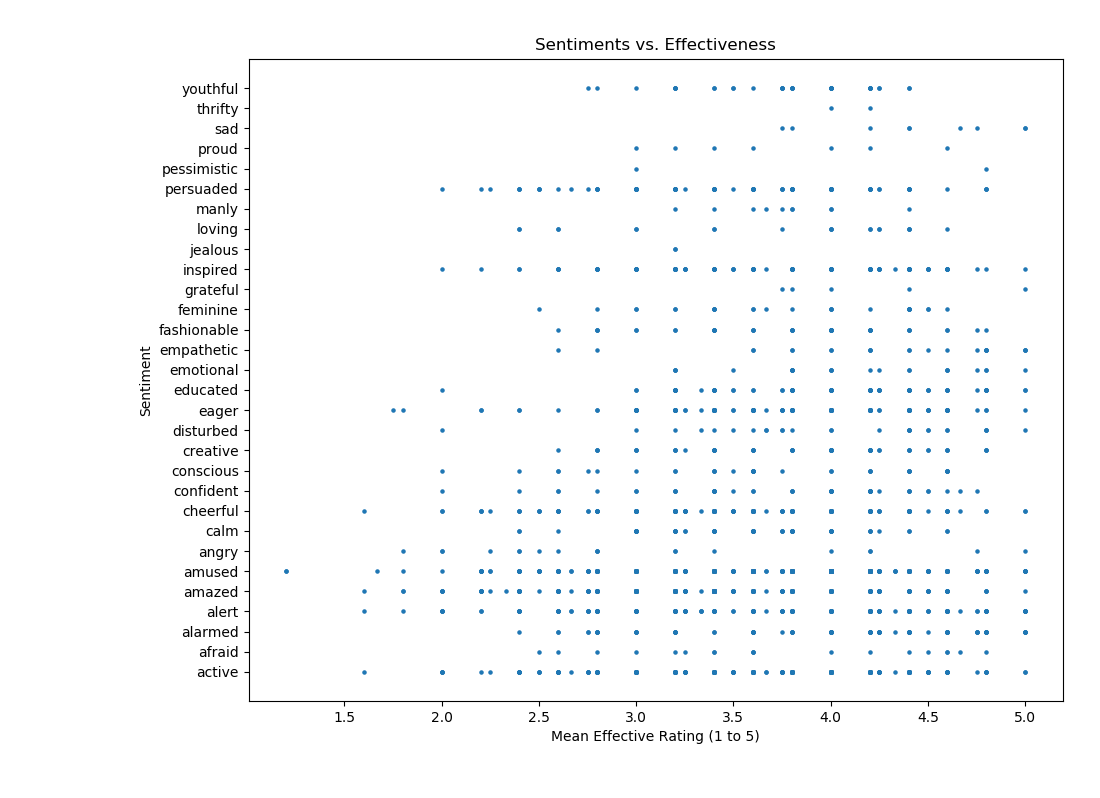}}
  \caption{Distribution of effectiveness ratings across each topic and sentiment}
  \label{dot-distributions}
\end{figure}

In search of useful indications of effectiveness, the 200 most effective and least effective advertisements were grouped by topic and sentiment.  Results can be seen in Figures \ref{topics-distributions-pie} \& \ref{sents-distributions-pie} respectively.  Keep in mind the dataset contains an uneven distribution of each topic and sentiment (e.g. 'safety' might show up three times more often than 'automobiles').  Therefore, if raw results were plotted in a pie chart, they may be skewed.  For example, if videos with topic 'safety' take up 5\% of the entire dataset and take up 5\% of the 200 most effective ads, this is to be expected.  This follows the ground truth distribution of the entire dataset.  Also, Figure \ref{sents-distribution-bar} represents how much more likely a given sentiment is to appear in the 200 most or least effective ads compared to the overall dataset, while Figure \ref{topics-distribution-bar} represents the same information for topics.  As supplementary analysis, topics' and sentiments' vs. effectiveness distributions are shown in Figure \ref{dot-distributions}.  Each data point, or dot, represents at least one sample with that mean effectiveness rating.

Finally, analysis of some videos in the dataset was performed by watching close to two hundred different videos, roughly 5\% of the population.  One video, \href{https://www.youtube.com/watch?v=jhFqSlvbKAM}{https://www.youtube.com/watch?v=jhFqSlvbKAM}, stands out since it contains Kobe Bryant and Lionel Messi, two of the most popular basketball and soccer players in the world.  Also, the video itself has 146m videos on YouTube.  The mode of annotator labels was '3' with individual ratings being ['5', '3', '3', '4', '5'], indicating celebrities are not guaranteed to make an ad effective.  As another example, \href{https://www.youtube.com/watch?v=-usbQDfTIqE}{https://www.youtube.com/watch?v=-usbQDfTIqE} is an Atari commercial from 1982.  Although the production quality is poor compared to modern advertisements, annotators listed the video as a '5' with individual ratings of ['4', '5', '5', '3', '4'], which comes out to an average of 4.2.  After averaging the previous video's ratings, coming out to a '4', the second video was deemed more effective than the first despite the presence of celebrities, higher production quality, more YouTube views, and more modernity.  Additionally, many videos associated with obesity, drugs, addictions, and related topics typically contained dark music, low video saturation, and sad facial expressions.  A final mention is the use of positive facial expressions with product placement.  For example, take \href{https://www.youtube.com/watch?v=2Md5lPyuvsk}{https://www.youtube.com/watch?v=2Md5lPyuvsk}, starring Michael Jackson.  Every scene change contains a can of Pepsi next to a happy child, furthering this claim.  In general, every video contains a diverse mix of features, not always leading to the most reasonable deductive conclusion in terms of effectiveness.  As shown, YouTube views do not always equate to higher effectiveness as mentioned earlier, and a plethora of features can be gathered from video processing to gain further insights into these advertisements' effective ratings.

\begin{algorithm}[h]{
$\text{// 25 arrays for the 25 different classifiers}$\\
$\text{Topics[25][38]} \gets \text{List of length 38 initialized to 0}$\;
$\text{Sentiments[25][30]} \gets \text{List of length 30 initialized to 0}$\;
$\text{Classifiers[25]} \gets \text{all 25 different classifiers to train}$\;
\For{$i \gets 1$ to $25$}{                    
    Classifiers[i].train(trainX, trainY)
}
\For{$n \gets 1$ to $N$}{                    
    \For{$i \gets 1$ to $25$}{     
        $\text{predictedClass} \gets \text{Classifiers[i].predict(trainX[n])\;}$\\
        $\text{videoTopic} \gets \text{video topic from 0 to 37\;}$\\
        $\text{videoSentiment} \gets \text{video sentiment from 0 to 30\;}$\\
        \If{predictedClass == trainY[n]}{
           Topics[i][n] += 1\;
           Sentiments[i][n] += 1\;
        }
    }
}
\For{$k \gets 1$ to $38$}{                    
    \For{$i \gets 1$ to $25$}{     
        Topics[i][k] /= (\# videos with this topic)\;
        Sentiments[i][k] /= (\# videos with this sent)\;
    }
}
\For{$m \gets 1$ to $M$}{
    $\text{topic} \gets \text{test video topic from 0 to 37\;}$\\
    $\text{sent} \gets \text{test video sentiment from 0 to 30\;}$\\
    $\text{c} \gets \argmax_{i}\text{[Topics[i][topic], Sentiments[i][sent]]\;}$\\
    $\text{predictedClass} \gets \text{Classifiers[c].predict(testX[m])\;}$\\
}
\caption{Hybrid Ensemble Learning Algorithm}
\label{ensemble-algo}
}\end{algorithm}

\subsection{Learning}

Training and testing classification algorithms on this data was simple, requiring anywhere from several seconds to several minutes.  As such, any attempt at boosting accuracy never had to deal with changing hardware on a machine or optimizing code, allowing quick investigation into key challenges of this task.

The first challenge to tackle was the dataset's size.  After preprocessing and feature gathering, 23 features spanned 594 dimensions.  With a dataset of 965 total samples, a 594 dimensional feature vector is an infeasible training vector.  Therefore, individual support vector machines (SVMs) \cite{c5}, decision trees \cite{c36}, and logistic regression \cite{c37} models were trained on each feature and a hybrid of bagging and stacking \cite{c6}, two common ensemble learning techniques, was used to aggregate class predictions of the classifiers.  Not only does this approach prevent overfitting, a serious concern with this data, but it also boosts confidence of the individually weak classifiers' predictions.

Specifically, each SVM was trained to achieve optimal results.  Accuracies in Table I were the average accuracies of each SVM trained on the respective feature across five simulations.  Beforehand, the dataset was randomly split for 80\% training and 20\% testing.  Each SVM was rigorously tested in terms of the hyperparameters.  For every classifier, the one-vs-rest multiclass algorithm was utilized to convert the naturally binary SVM into a multiclass classifier.  This almost always produced superior results compared to one-vs-one, as well as reduced training and testing time.  Different kernels worked better for different SVMs, which was discovered after rigorous testing, but most classifiers used either a polynomial or linear kernel.  In Table I, results for binary classification, four-way classification, and five-way classification across the effectiveness ratings are shown.  Neural networks were avoided due to their requirement of large datasets and greater risk of overfitting.

\begin{figure}[t]
  \centering
  {\label{ref_label1}\includegraphics[width=200px,keepaspectratio]{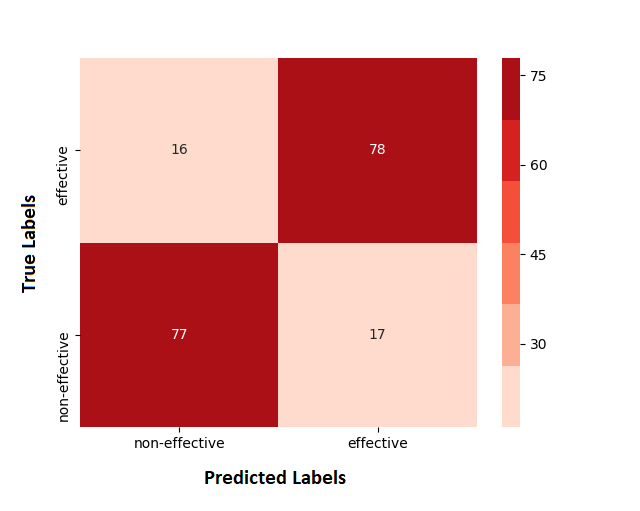}}
  {\label{ref_label2}\includegraphics[width=200px,keepaspectratio]{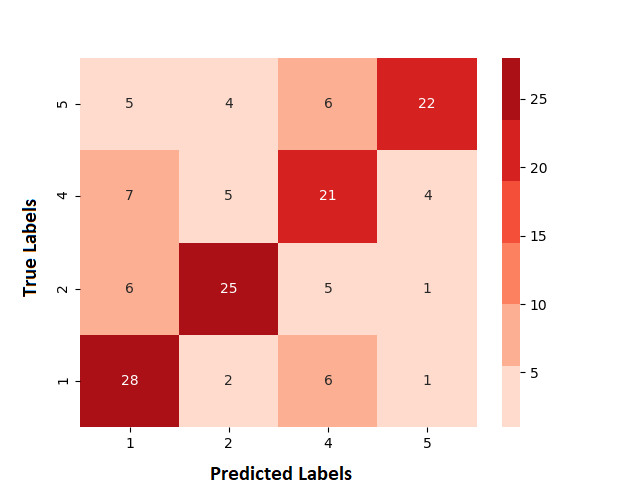}}
  {\label{ref_label2}\includegraphics[width=200px,keepaspectratio]{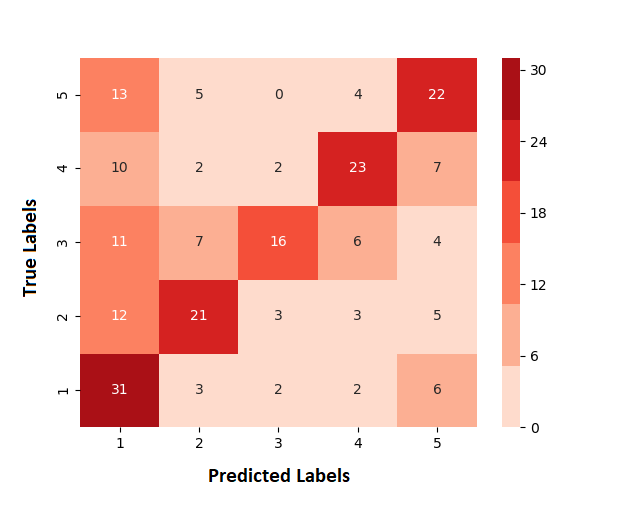}}
  \caption{Confusion Matrix of each multiclass classification task}
  \label{confusion-matrices}
\end{figure}

In addition to the SVMs, a logistic regression classifier was trained on the 'exciting' feature and a decision tree was trained on the topics and sentiments features.  By default, a logistic regression model in sci-kit learn utilizes one-vs-rest for multi-class classification.  Probabilities of classification were not utilized as a weighting parameter in terms of accuracy or further analysis.  A decision tree was investigated for the topics and sentiments features as test runs to see if they boosted accuracy.  Both tree classifiers were trained using the Gini impurity metric (default on sci-kit learn and standard in the scientific community) and new branches were created until a depth with a minimum splitting criterion was reached.

The unique hybrid ensemble learning algorithm, shown in Algorithm \ref{ensemble-algo}, provides the most significant boost in accuracy to predictions.  To implement this, all individual classifiers were first trained on 80\% of the dataset.  Then, for each classifier, the accuracy was computed for each topic and for each sentiment on the training data, resulting in 68 `bins' of accuracies.  To classify the testing dataset, the classifier with the highest accuracy for that test video's topic or sentiment was chosen to predict the output label.  In this sense, a form of stacking was utilized since each classifier had a 'weight' assigned to it (its accuracy on test video's topic or sentiment).  It is also a form of bagging since each classifier votes on a class for each iteration.  Only the class with the largest vote dictates the final prediction. Pseudocode of this hybrid approach is represented in Algorithm 1.

\subsection{Experiments}

As shown, the hybrid ensemble learning algorithm produces accuracies far surpassing the baseline accuracies of 20\% on five-way classification, 25\% on four-way classification, and 50\% on binary classification.  In addition, these accuracies were achieved on small dataset of merely 965 samples.  This is a significant feat since many modern machine learning and computer vision tasks require several magnitudes of samples higher in order to accomplish notable results.

Of the features, the highest in terms of accuracy was the advertisement's topic, duration, exciticement level.  Also, the classifier trained on all features stood out noticeably with a 56.69\% binary accuracy.  On four-way classification, results were relativey the same with a few more standout features, such as the ad's sentiment and average hue.  Finally, most classifiers on five-way classification performed close to the baseline, but the average hue, human emotions, all features aggregated, sentiments, and exciting classifiers produced the best results, deviating significantly from the 20\% $\pm$ 1\% accuracies.

In Figure \ref{confusion-matrices}, the three confusion matrices displaying true positives, false positives, true negatives, and false negatives are shown for all three classification tasks (five-way, four-way, binary).  The binary task matrix is straightforward.  For any misclassified sample, it is simply placed into the other false positive or false negative bin.  Thankfully, the true positive and true negative bins contain similar numbers, so there are no significant gaps in misclassification errors.  For the four-way task, most classifications were true positives.  Bins with misclassified samples largely remained in the 4-6 sample range.  Statistically, there are no misclassification bin outliers; no misclassified bin contains a significant portion of that class's samples compared to other bins.  Results are more ambiguous with the five-way classification, as to be expected.  True positive results are consistent; effectiveness 1 and 5 have highest true positive ratios.  Meanwhile, effectiveness rating 3 is the most ambiguous.  Clearly, with the fewest true positive ratings, many videos were rated as 2 or 4.  This issue did not arise with other effectiveness ratings, other than a medium amount of samples from effectiveness 4 being predicted as effectiveness 5.  Interestingly, a large number of samples from each class were predicted as effectiveness 1.  This includes both classes 4 and 5, which are the most shocking.  Investigation into the machine learning models, for example the support vectors, does not provide insight into the reasoning why many samples are misclassified this way.

In general, most samples were classified correctly and appeared as true positives.  This is to be expected from the achieved accuracies.  Furthermore, in addition to successful prediction of effectiveness ratings, statistical inference was provided for further insights into misclassifications, specific advertisement analysis, and individual feature classifier accuracies showcasing its success, rather than aggregating all features together as a blackbox.

\section{Conclusions}

As shown in Figure \ref{accuracies} and discussed previously, no feature was a significant indicator of effectiveness by itself.  But, our ensemble proved effective, boosting accuracy on the tasks by as much as 35\%.  Clearly, an ensemble of weak features on this dataset provides an accurate model of predicting advertisement effectiveness.  

We presented a unique approach computational approach on a new advertisements dataset which has not been seen before in the fields of media studies and computer vision.  This approach arose from simple human annotated labels, with most of them being metafeatures of the video itself, such as language, topic, and sentiment, while others were extracted through common video processing techniques.  Finally, features from an additional paper were brought in to showcase the usefulness of higher level characteristics, such as climaxes, objects, and facial expressions.  This standalone approach requires minimal human input, if any, relying heavily on computational aspects, which many marketing companies and media studies papers do not have in place.

With that being said, there is an abundance of future work in the the field of computional advertisement understanding, and specifically in measuring effectiveness.  In the related work, other papers explored additional features, such as celebrities, from a non-computational perspective, which can be brought into the computational sphere.  This work also has potential create a pipeline for advertisement generation in the future.  Finally, if a larger dataset is gathered, neural networks become more of a possibility, as they require large datasets to be trained effectively.  There are several avenues to explore future work in this fiel

\section{Future Work}
Through this research, we have displayed feasibility of measuring effectiveness of advertisements through various audio, visual, and metadata features of a given video, excluding the use of YouTube comments or ratings, which are unavailable for a video before it is officially released to the public.  

One future avenue of work is exploration of the static image dataset.  The most significant challenge with the video dataset is the initially small sample size of 3,477 videos, which was further reduced to 965 samples after normalizing across effectiveness.  The image dataset contains $\approx$ 64k images.  Therefore, decreased risk of overfitting and more accurate computation of a ground-truth prior distribution for classifiers should result from the use of this dataset.  It will also allow use of a validation set, many-fold cross-validation, and data modelling with higher complexity.  Furthermore, the task will generally be less arduous.  Videos contain more information than static images, such as duration, optical flow data, shot boundary data, climax data, and more.  Despite this lack of information, static images should be more straightforward in which features to extract, resulting in higher clarity of results of the advertisement's faults.

In terms of harvesting additional information, the idea of celebrity detection was initially thought useful since popular clothing and product brands (i.e. Adidas for Lionel Messi and Nike for Lebron James) sign multi-million sponsorship deals \cite{c38}.  Surely, this implies these brands value a celebrity's popularity to gain additional revenue for their brand.  Also, GumGum \cite{c39}, a computer vision startup based in Los Angeles, provides valuations of brands on specific sports team's jersey's, further supporting this hypothesis.  However, due to monetary concerns, this feature was not explored.

A unique approach with potential to boost accuracy is training a neural network, such as a convolutional neural network, to predict the topic or sentiment of a video.  In turn, this showcases how well a computer is able to represent a video's overall theme.  A higher accuracy indicates an advertisement's message is successfully transferred to a human audience, increasing effectiveness.

Recently, there has been research on infographics, a form of modern advertisement with heavy information content \cite{c34}.  Bylinskii et al. investigated textual information flow and predicted tags suitable for the content in the infographic.  This same idea can be applied to advertisements, but moreso with static images rather than video advertisements, to place tags on internet advertisements on social media sites, such as Facebook, Instagram, and YouTube, further enhancing effectiveness and outreach of a brand or product.

A feat with significant capital potential is constructing a formula for effective video advertisements.  For example, generating permutations of features and testing their individual effectiveness, then ranking their effectiveness.  Furthermore, this ranking can become more granular for specific demographics, such as kids, adults, cultures, geographic regions, and income levels. To supplement this idea, a tool to recommend improvements to an advertisement can help boost additional revenue or outreach initiatives before the advertisement is released to the public via the internet, television, or mobile devices.

As a notable recent technological advance, Burger King has experimented with the use of artificial intelligence to generate advertisement descriptions for their products \cite{c10}.  Once generated, they utilize text-to-speech software to annotate their advertisements.  A potential future application is automatic generation of video advertisements with the use of generative adversarial networks \cite{c9}.  Paired with the aforementioned improvement recommendation tool, a challenge is posed to generate advertisements, provide feedback to itself, and create highly effective videos.  This task is difficult due to lack of training data, high complexity of video advertisement features, and recent developments of GANs.

Finally, although the dataset provides a thorough insight into the world of advertisements, not all videos were from major brands.  Therefore, a slight bias may have been introduced into the dataset.  It would be interesting to investigate the comparison of these results with the results from a commercial advertisement dataset, perhaps from a major cable television network or popular streaming service.

As a result, this research has opened up several gateways to boosting how content creators generate advertisements and how consumers consume advertisements.  Additionally, several avenues of research have been proposed to generate momentum within the computer vision advertisements community with new, interesting challenges.

\bibliographystyle{plain}
\bibliography{bibliography.bib}

\end{document}